\documentclass[preprint,5p,twocolumn,authoryear]{elsarticle}

\usepackage{amssymb}
\usepackage{amsmath}
\usepackage{multirow}

\journal{Medical Image Analysis}

\begin{document}

\begin{frontmatter}

\title{Adversarial Versus Federated: An Adversarial Learning based Multi-Modality Cross-Domain Federated Medical Segmentation}

\author[Beihang]{You Zhou} 
\ead{sy2402322@buaa.edu.cn}
\author[Beihang]{Lijiang Chen}
\ead{chenlijiang@buaa.edu.cn}
\author[Beihang]{Shuchang Lyu\corref{corresponding}}
\ead{lyushuchang@buaa.edu.cn}
\author[shijitan]{Guangxia Cui}
\ead{cuigx@bjsjth.cn}
\author[shijitan]{Wenpei Bai}
\ead{baiwp@bjsjth.cn}
\author[Beihang]{Zheng Zhou}
\ead{zhengzhou@buaa.edu.cn}
\author[Beihang]{Meng Li}
\ead{limenglm@buaa.edu.cn}
\author[liverpool]{Guangliang Cheng}
\ead{Guangliang.Cheng@liverpool.ac.uk}
\author[Leicester]{Huiyu Zhou}
\ead{hz143@leicester.ac.uk}
\author[Beihang]{Qi Zhao}
\ead{zhaoqi@buaa.edu.cn}
\affiliation[Beihang]{organization={Department of Electronics and Information Engineering, Beihang University},            
addressline={Xueyuan Road No.37}, 
            city={Haidian district},
            postcode={100191}, 
            state={Beijing},
            country={China}}
\affiliation[shijitan]{organization={Department of Gynecology and Obstetrics, Beijing Shijitan Hospital, Capital Medical University},    
addressline={Tieyi Road No.10}, 
            city={Haidian district},
            postcode={100038}, 
            state={Beijing},
            country={China}}
\affiliation[liverpool]{organization={Department of Computer Science, University of Liverpool},            
addressline={Foundation Building, Brownlow Hill}, 
            city={Liverpool},
            postcode={L693BX},
            country={UK}}
\affiliation[Leicester]{organization={School of Computing and Mathematical Sciences, University of Leicester},            
addressline={Room 529B, Ken Edwards Building}, 
            city={Leicester},
            postcode={LE17RH},
            country={UK}}
            
\cortext[corresponding]{Corresponding authors}
\begin{abstract}
Federated learning enables collaborative training of machine learning models among different clients while ensuring data privacy, emerging as the mainstream for breaking data silos in the healthcare domain. However, the imbalance of medical resources, data corruption or improper data preservation may lead to a situation where different clients possess medical images of different modality. This heterogeneity poses a significant challenge for cross-domain medical image segmentation within the federated learning framework. To address this challenge, we propose a new Federated Domain Adaptation (FedDA) segmentation training framework. Specifically, we propose a feature-level adversarial learning among clients by aligning feature maps across clients through embedding an adversarial training mechanism. This design can enhance the model's generalization on multiple domains and alleviate the negative impact from domain-shift. Comprehensive experiments on three medical image datasets demonstrate that our proposed FedDA substantially achieves cross-domain federated aggregation, endowing single modality client with cross-modality processing capabilities, and consistently delivers robust performance compared to state-of-the-art federated aggregation algorithms in objective and subjective assessment. Our code are available at https://github.com/GGbond-study/FedDA.
\end{abstract}


\end{frontmatter}

\section{Introduction}
\begin{figure}
    \centering
    \includegraphics[width=\linewidth]{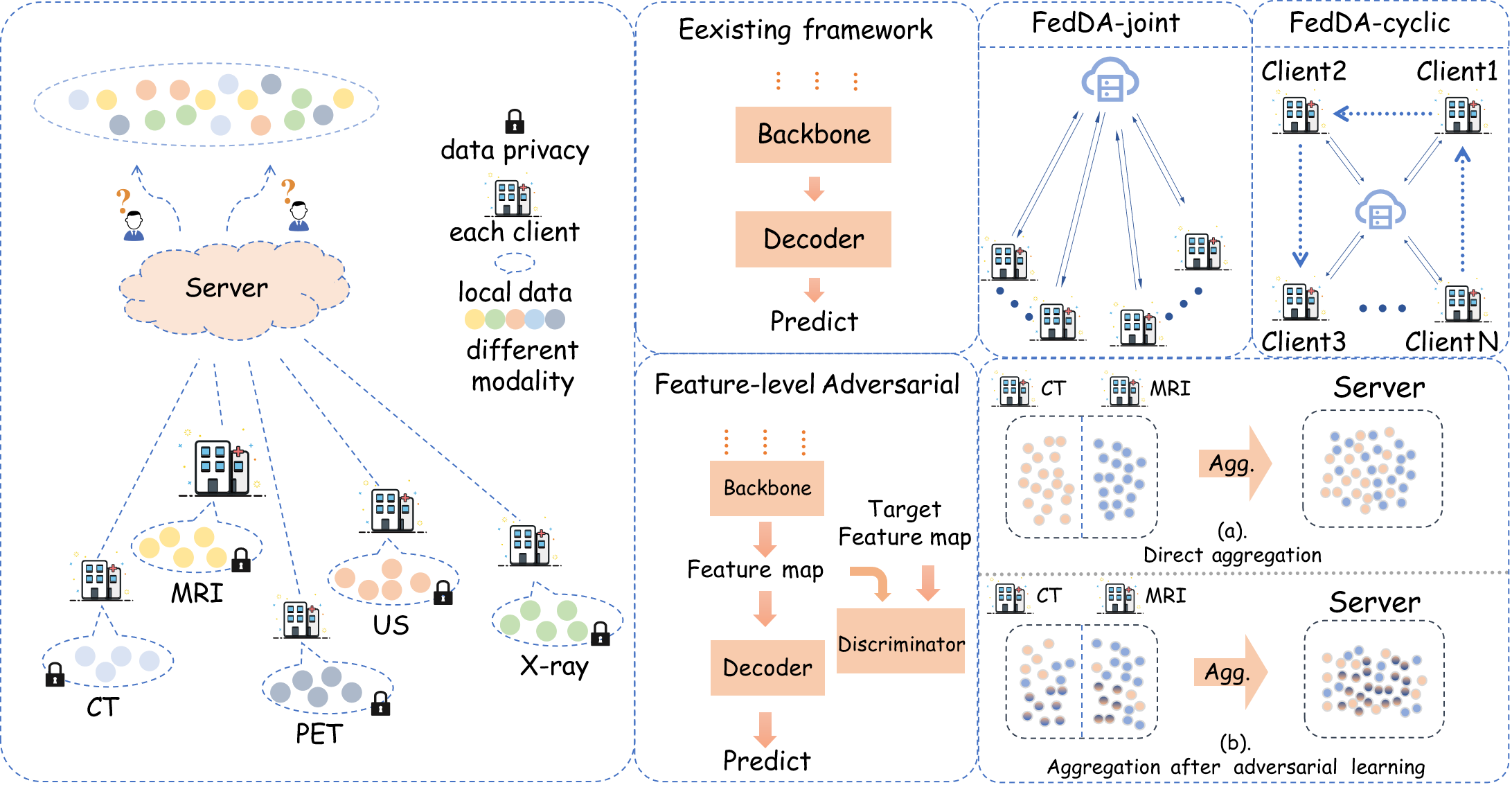}
    \caption{Compared to traditional federated learning algorithms, we introduced adversarial training between source and target features by transmitting them among clients. After incorporating adversarial learning, the model aggregation among clients is no longer a simple averaging process but rather a mutual fusion.}
    \label{flow}
\end{figure}

Deep learning models are inherently data-dependent, requiring large-scale and representative datasets to achieve optimal performance~\citep{li2021model,ma2024fedst}. However, medical data are typically distributed among various medical institutions and research centers due to privacy concerns~\citep{ma2024fedst, lanfredi2025enhancing}. Given the premise of preserving data privacy, data collaboration training across multiple medical institutions is increasingly desired to build accurate and robust data-driven deep networks for medical image segmentation~\citep{chen2025sam, dang2025cad, shilo2020axes,kaissis2020secure, gu2025dual}. Federated learning (FL) has recently opened the door for a promising privacy-preserving solution, which enables multiple parties to collaborate on model development without sharing raw data~\citep{wu2024feda3i}. The paradigm works in a way that each local client (hospital) learns from their own data, and only aggregates the model parameters at a certain frequency at the central server to generate a global model~\citep{liu2021feddg}. Despite the potential of federated learning to address data privacy concerns, the inherent heterogeneity of medical data across different institutions presents additional hurdles. For instance, an imbalance in medical resources distribution or failures in data management such as corruption or improper preservation can create a scenario where clients (e.g., hospitals or clinics) are left with fragmented datasets. As shown in Fig.~\ref{flow}, some clients may only possess medical images of one modality (e.g., MRI), while other clients hold images of a different modality (e.g., CT). This non-typical non-IID (non-independently and identically distributed) federated learning scenario poses significant challenges, resulting in worse performance and slower convergence rates. Recent research in federated medical segmentation~\citep{jiang2025shadow, liu2025federated, zhu2025fedbm, bitarafan2025self, chen2025label,chen2025sam} has primarily focused on fairness~\citep{ezzeldin2023fairfed}, continual learning~\citep{wang2024traceable}, label imbalance~\citep{bai2024combating}, temporal shifts~\citep{derakhshani2022lifelonger}. How to generalize the federated model under such distribution shifts is technically challenging yet unexplored so far. In this work, we identify a new problem setting of Federated Domain Adaptation (FedDA), which aims to learn a federated model from multiple decentralized source domains in order to achieve excellent performance across all modalities on all clients (Fig.~\ref{flow}).
\par To address this non-typical non-IID problem, we tackle this challenge from a new perspective based on an intuitive observation: \textit{the model trained on each local dataset is able to exhibit the representation of its own data through high-dimensional, complex and irreversible feature maps.}
Based on this observation, we propose a new Federated Domain Adaptation (FedDA) segmentation training framework that introduces a feature-level adversarial training to correct the local updates by maximizing the agreement of representation learned by the current local model and the representation of models across all clients. Thus each client can be guided by this feature-level adversarial during each local update toward consistent feature representations which alleviates the influence of non-IID phenomenon.
\par Unlike typical domain adaptation approaches~\citep{ganin2016domain} that perform feature-level alignment by comparing representations between two domains, our proposed federated adversarial framework is not limited to pairwise interactions but instead elevates to a global and cyclic adversarial mechanism across multiple clients. As shown in Fig.~\ref{flow}, we introduce FedDA with two distinct modes: FedDA-cyclic and FedDA-joint, depending on whether the feature maps required for local adversarial training are generated by the global model or transferred from other clients. The FedDA-cyclic introduces a cyclic client-selection mechanism, where the learned representations are sequentially passed among clients in a cyclic manner, enabling progressive adversarial learning across clients. The FedDA-joint conducts adversarial training by aligning the global model with local models using feature maps generated from local data during each training round, ensuring continuous alignment throughout iterative updates.
\par These two variations both enable adversarial alignment and progressive fusion of local feature representations between multiple clients during federated training. After incorporating adversarial learning, the model aggregation among clients is no longer a simple averaging process but rather a mutual fusion. After integrating the features from different clients, although the modality differences still exist, the aggregation process reduces the associated losses in each round, thereby enhancing the generalization ability of the global model. In addition, while FedDA-cyclic involves transferring feature maps of local clients among clients which may raise privacy concerns,  feature map of each client only partially reflects the representation of local data. As high-dimensional, complex, and non-invertible tensors, these feature maps do not reveal specific information about the original data, thus it has little impact on the privacy of each client.
In general, our main contributions are highlighted as follows:
\begin{itemize}
    \item We propose and formalize a practical and challenging problem of non-typical non-IID federated domain adaptation, where clients exhibit significant modality heterogeneity. To the best of our knowledge, this is the first work to systematically address the scenario of modality differentiation across clients in federated learning.
    \item To mitigate the domain shift problem present in clients possessing single-modality data, we propose FedDA, a federated domain adaptation framework leveraging feature-level adversarial learning for cross-client representation alignment.
    \item We present two distinct FedDA variants— FedDA-cyclic and FedDA-joint —differentiated by their feature map sources for local adversarial training designed to explicitly mitigate client heterogeneity.
    \item Extensive experiments on three international medical datasets validate that FedDA outperforms state-of-the-art methods in both subjective and objective evaluations.
\end{itemize}
\begin{figure*}
    \centering    
    \includegraphics[width=\linewidth]{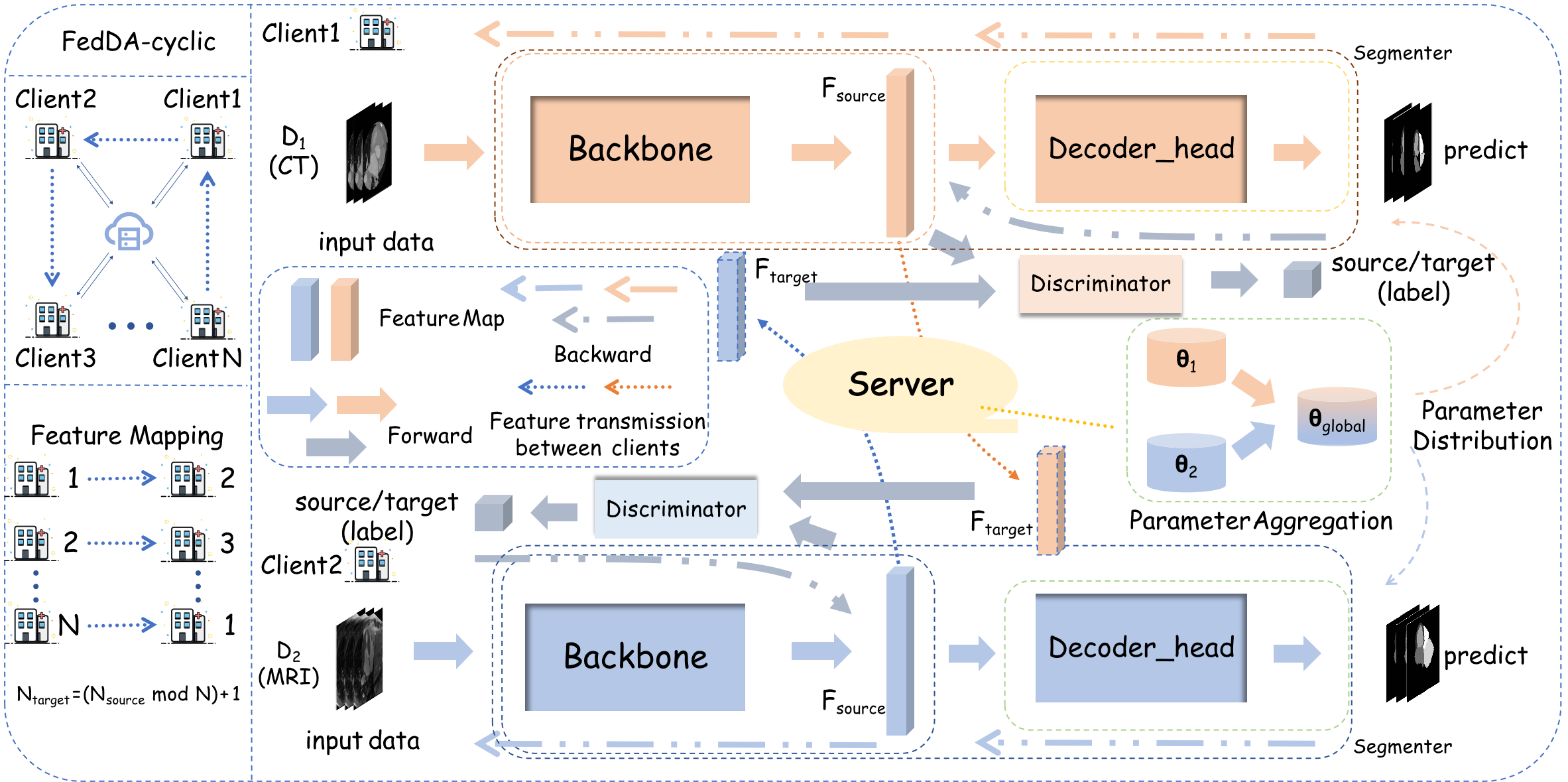}    
    \caption{Overview of the proposed FedDA-cyclic. After each local training, the clients upload model parameters and feature maps to the server. Based on the feature mapping, each client will use the mapped feature maps as the target features for its local training. In the scenario involving two clients, we exchange the feature maps between clients and conduct adversarial training between the source and target feature maps.}    
    \label{frame}
\end{figure*}
\section{Related Work}
\subsection{Federated Medical Image Segmentation}
Federated learning~\citep{konevcny2016federated, yang2019federated,halpern2025federated} provides a promising privacy-preserving solution for decentralized data collaboration by developing a global model from decentralized datasets through aggregating local client parameters while keeping data locally. In the federated learning framework, clients (which can range from mobile devices and edge servers to enterprise data centers) independently train machine learning models on their local datasets using algorithms such as gradient descent or stochastic optimization techniques. Instead of sending raw data to a central server, clients only transmit model updates (e.g., gradients or weights) to a parameter server. This server then aggregates these updates using sophisticated aggregation algorithms. McMahan et al.~\citep{mcmahan2017communication} first introduced the Federated Averaging algorithm (FedAvg), enabling communication-efficient federated training of deep networks, which has since established a foundational paradigm for federated learning. After that, FedProx~\citep{li2020federated} introduces a proximal term into the objective function during local training. Then, Li et al.~\citep{lifedbn} prove that excluding batch normalization layers from parameter aggregation accelerates convergence and improves model performance in federated learning scenarios. In the same year, MOON~\citep{li2021model} employs model-contrastive learning at the model-level to guide the update direction of local models. Shortly afterward, FedMLB~\citep{kim2022multi} employs knowledge distillation to construct auxiliary branches, thereby mitigating the adverse effects caused by data heterogeneity. Latest literature~\citep{wu2024feda3i, zhou2024federated} have studied unsupervised federated learning or scenarios with low-quality labels. Instead, our work mainly focuses on how to achieve excellent performance across all clients when substantial modality heterogeneity exists among different clients.
\begin{figure*}
    \centering    
    \includegraphics[width=\linewidth]{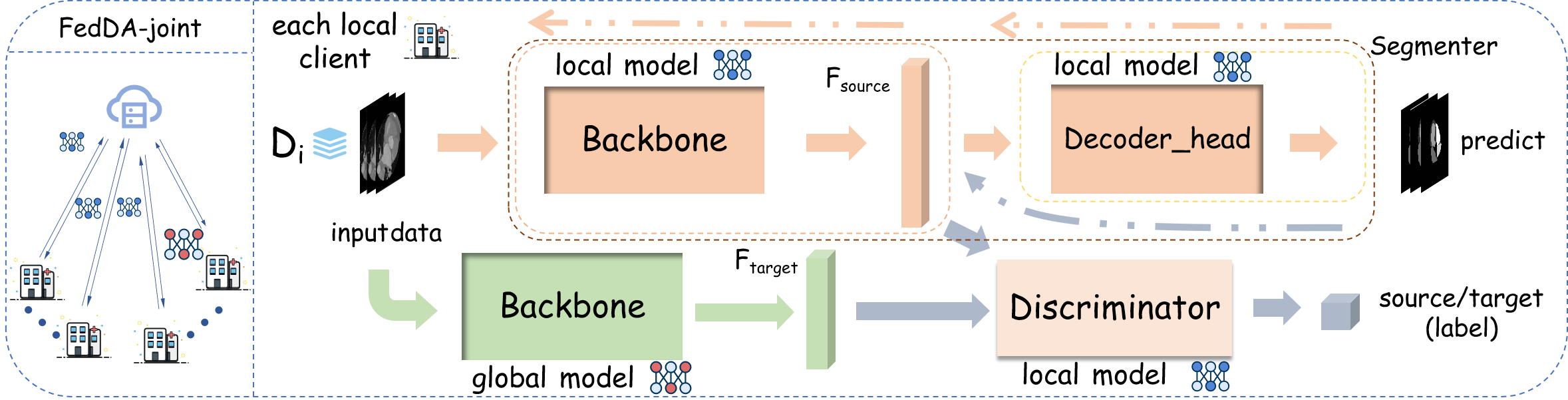}  
    \caption{Overview of the proposed FedDA-joint. Upon receiving the global parameters, each client will perform feature-level adversarial training between the local model and the global model.}    
    \label{joint}
\end{figure*}
\subsection{Domain Adaption in Medical Imaging}
In the realm of machine learning, datasets often exhibit variations in characteristics such as data distribution, feature representation, and label space across different domains~\citep{tang2024source}. These discrepancies~\citep{broni2024unsupervised} can lead to significant performance degradation when a model trained on a source domain is directly applied to a target domain, a phenomenon known as the domain shift problem. Domain adaptation aims to transfer a model from one domain (source domain) to another different domain (target domain), thereby improving the model's performance on the target domain~\citep{ganin2016domain}. Men et al.~\citep{men2023towards} use domain adversarial training to map features from source-domain videos to target-domain for image quality. Fan et al.~\citep{fan2024domain} propose an attention based deep convolution neural network with domain adaption to ameliorate seizure diagnosis. Judge et al.~\citep{judge2024domain} introduce a reinforcement learning framework that reduces the need to otherwise incorporate large expertly annotated datasets in the target domain, effectively alleviating challenges in domain adaptation. These methods~\citep{ zhu2025improving,zhang2025enhancing} assume the data is centralized on one server, limiting their
applicability to the distributed learning system. Moreover, in light of the stringent data privacy constraints inherent in federated scenarios, it is extremely difficult to match the statistics of distributions~\citep{wang2022personalizing}. 
Therefore, our work focuses on employing adversarial-based domain adaptation to align the features of each client model, thereby reducing the loss incurred during federated model aggregation.
\section{Methodology}
\subsection{Problem Definition}
Suppose there are $N$ parties, denoted as $P_1,\ldots,P_N$, where each $P_i$ ($1\leq i\leq N$) has a local dataset $\mathcal{D}_i$. The modality $m_i$ of dataset $\mathcal{D}_i$ belongs to a distinct medical imaging type (e.g., CT, MRI, etc.), varying across different clients. Some clients possess medical images of one modality, while other clients hold images of a different modality, shown in the Equ~\ref{sit}.
\begin{equation}
\exists i,j \in (0,N),m_i\cap m_j=0
\label{sit}
\end{equation}
Our goal is to train a global model $w$ to achieve excellent performance on the all modality ($m \in  {\textstyle \bigcup_{i\in[0,N]}}m_i $) of all dataset ($\mathcal{D} =  {\textstyle \bigcup_{i\in[0,N]}} \mathcal{D}^i$). The objective is to solve 
\begin{equation}
\arg \min_w \mathcal{L}(w) = \sum_{i=1}^N \frac{|\mathcal{D}_i|}{|\mathcal{D}|} L_i(w)
\label{goal}
\end{equation}
where \( L_i(w) = {E}_{(x,y) \sim \mathcal{D}_i} [\ell_i(w; (x, y))] \) is the loss of \( P_i \) and $|\mathcal{D}|$ represents the size of dataset $|\mathcal{D}|$. Note that exchanges of training data are strictly prohibited due to privacy issues.
\subsection{Network Architecture}
To address the challenges posed by this non-typical non-IID federated learning scenarios, we propose a new federated framework called FedDA which has three components: a backbone $B$, a decoder $H$ and  a feature discriminator $D$. In the federated domain adaptation scenarios, from the perspective of each local client, we can regard the feature maps generated by the local model on the local data as the source feature maps. Meanwhile, the feature maps originating from other clients are defined as target features maps. Taking two-client scenario as an illustrative case (Fig.~\ref{frame} and Fig.~\ref{joint}), the feature map $\{F_{k}^{i}\}$ is generated by the $i$-th iteration of the $k-th$ client. In the two-client scenario, the source feature map of $Client1$ corresponds to the target feature map of $Client2$. 
\par We train the discriminator to maximize the probability of correctly assigning labels to both the target features and the source features extracted by the backbone. Consequently, the backbone is encouraged to generate feature maps that are more aligned with the underlying distribution patterns. Specifically, the backbone will be updated to minimize the differences between its own feature maps and those of other clients, while also optimizing towards the local segmentation objectives.
\subsection{From FedDA-cyclic to FedDA-joint}
The distinction between FedDA-cyclic and FedDA-joint lies in whether the target feature maps on each local client are generated by the global model or transmitted from other clients. In FedDA-cyclic, each client separately generates their own source feature maps using their local domain image and local model, uploads it to the server and saves it to the server feature bank. Then, in the subsequent federated learning rounds, each client will obtain its corresponding feature map according to a specific mapping, and use it as the target feature for local training. The specific mapping is formally defined by the subsequent equation (shown in Equ.~\ref{map}), resembling a cyclically ordered client ensemble. 
\begin{equation}
    N_{target} = (N_{source} \mod N) + 1
    \label{map}
\end{equation}
where \( N_{{source}} \) and \( N_{{target}} \) represent the indices of the current and target clients, respectively. During each round of federated training, each client generates local feature maps and stores them in the server feature bank. Subsequently, the client retrieves the corresponding target feature maps for local adversarial training, thereby reducing the representation distance among different clients. 
\par
In fact, safeguarding client privacy and mitigating communication overhead constitute two key challenges in federated learning, especially when dealing with non-IID data across clients. To mitigate these issues, we further propose the FedDA-joint method. \textbf{This method introduces no additional communication overhead and ensures no leakage of any information.} After each local client receives the global model, the feature map generated by the global model on the local data is regarded as the target feature map. During each iteration of training, the feature map produced by the local model on the local data is referred to as the source feature map. During the local adversarial training between the target feature map and the source feature map, each training iteration continuously updates the local segmentation model in a direction that aligns with both the global model and the improved local segmentation performance. It can reduce the associated communication overhead compared with FedDA-cyclic and alleviates the impact of non-IID phenomenon.
\begin{table}[]    
\centering    
\fontsize{9pt}{10pt}\selectfont    
\setlength{\tabcolsep}{0.52mm}     
    \scalebox{0.752}{        
    \begin{tabular}{c|cccccc|cccccc}        
    \hline    
    \multirow{3}{*}{Methods} & \multicolumn{12}{c}{Cardiac Substructure Segmentation (MMWHS)} \\ \cline{2-13}    
    & \multicolumn{6}{c}{Dice (\%) $\uparrow$} & \multicolumn{6}{c}{HD95$\downarrow$}  \\ \cline{2-13}    
    & MYO & LA & LV & RA & RV  & Mean & MYO & LA & LV & RA & RV & Mean\\ \hline    
    FedAvg & 61.2 & 83.8 & 87.2 & 82.7 & 86.3 & 80.2 & 26.8 & 28.7 & 23.3 & 24.0 & 20.8 & 24.7 \\    
    Krum & 47.9 & 74.8 & 79.2 & 66.8 & 85.0 & 70.8 & 41.5 & 38.3 & 41.4 & 60.9 & 22.3 & 40.9\\    
    Fedprox & 64.3 & 83.2 & 87.8 & 85.1 & \underline{88.9} & 81.9 & 27.5 & 30.8 & 24.7 & 20.0 & 18.4 & 24.3\\    
    FedBN & 63.6 & 83.2 & 87.3 & 84.8 & 86.3 & 81.1 & 27.2 & 30.3 & 22.8 & 21.8 & 21.2 & 24.7\\
    Moon & 60.9 & 85.6 & 86.8 & 83.9 & 87.1 & 80.8 & 25.2 & 26.8 & 23.4 & 21.9 & 20.2 & 23.5\\    
    FedMLB & 65.2 & 85.9 & \textbf{88.9} & 83.1 & 87.8 & 82.2 & 25.9 & 28.4 & 23.2 & 31.8 & 18.6 & 25.6\\ \hline    
    FedDA-cyclic & \underline{67.6} & \underline{88.1} & 88.8 & \underline{88.3} & 88.9 & \underline{84.3} & \underline{21.7} & \underline{25.0} & \textbf{20.4} & \underline{16.5} & \underline{17.7} & \underline{20.3}\\    
    FedDA-joint & \textbf{68.2} & \textbf{88.3} & \textbf{88.9} & \textbf{88.8} & \textbf{89.7} & \textbf{84.8} & \textbf{21.3} & \textbf{22.4} & \textbf{20.4} & \textbf{15.3} & \textbf{16.9} & \textbf{19.3}\\ \hline        
    \end{tabular}
}    
\caption{The performance comparison of FedDA and other state-of-the-art methods on the two-client setting for the two medical image segmentation tasks. The bold value and underline value denote the best and the second-best performance in each metric, respectively. }    
\label{2client_m}
\end{table}
\subsection{Cross-Client Adversarial Training}
We train discriminator to maximize the probability of assigning the correct label to both target feature and source feature from backbone. We expect that backbone can generate feature maps that are closer to the distribution patterns. The adversarial relationship between the backbone and the discriminator during local training as follow: 
\begin{equation}
\begin{aligned}
    \min_B \max_D V(B, D) = {E}_{{x_t} \sim X_{t}} \left[ \log D(B(x_t)) \right] \\
    + {E}_{{x_s} \sim X_{s}} \left[ \log(1 - D(B(x_s)) \right]\\
\end{aligned}
\label{3}
\end{equation}
where \textit{B}, \textit{D} represents the backbone and discriminator of the same client. During the model training, gradient updates and backpropagation can be broadly divided into two components: Decoder \textit{H} and Discriminator \textit{D}. We use the cross-entropy loss to measure the error of label prediction on the source domain data and the binary cross-entropy loss for discriminator. The backbone $B_i$ aims to generate feature that can fool the discriminator $D_i$. Training the Discriminator improves its ability to clearly distinguish between locally generated feature maps and the target feature maps. The discriminator classifies the source feature maps $F_s$ as 0 and the target feature maps $F_t$ as 1, thereby differentiating between the two using discrimination loss $\mathcal{L}_D$:
\begin{equation}    
\begin{aligned}
\mathcal{L}_D = \frac{1}{n_s + n_t}  [\sum L_{BCE}(D_i({F}_{s}), 0) \\
+ \sum L_{BCE}(D_i({F}_{t}), 1)]  \\
\end{aligned}
\end{equation}
with ($n_s$,$n_t$) being the sample size of source image and target image domains and $L_{BCE}$ represents the binary cross-entropy loss.
\subsection{Model Parameter Aggregation}
In FedDA, only the parameters of the decoder and backbone are aggregated across different clients (Eq.~\ref{agg}). Due to the difference in the target feature maps for each client, the parameters of the Discriminator are not aggregated. 
\begin{equation}    
\theta = [\frac{\sum_{k=1}^{K} w_k \theta_k^{S} }{\sum_{k=1}^{K} w_k} ] + \theta^D
\label{agg}
\end{equation}
The whole parameters of each client $\theta$ are derived through the aggregation ${\sum_{k=1}^{K} w_k \theta_k^{S} }$ of the segmentation parameters associated with each client, followed by the Discriminator parameters $\theta^D$.
\begin{table}[t]
    \centering
    \fontsize{9pt}{10pt}\selectfont
    \setlength{\tabcolsep}{0.8mm} 
    \scalebox{0.825}{
        \begin{tabular}{c|ccccc|ccccc}
        \hline
    \multirow{3}{*}{Methods} & \multicolumn{10}{c}{Abdominal Multi-Organ Segmentation (CHAOS)}\\ \cline{2-11}
    & \multicolumn{5}{c}{Dice (\%) $\uparrow$} & \multicolumn{5}{c}{HD95$\downarrow$} \\ \cline{2-11}
    & LK & SP & LR & RK & Mean & LK & SP & LR & RK & Mean\\ \hline
    FedAvg & 54.6 & 44.1 & 72.6 & 57.1 & 57.1 & 14.9 & \textbf{9.9} & 69.7 & 22.5 & 29.2 \\
    Krum & 53.1 & 50.7 & 71.2 & 49.7 & 56.2 & \underline{13.0} & 10.9 & 78.7 & \textbf{15.7} & 29.6\\
    Fedprox & 55.0 & 49.6 & 76.5 & 58.9 & 60.0 & 15.3 & 13.3 & \underline{53.0} & 21.6 & 26.0\\
    FedBN & 52.7 & 52.1 & 74.5 & 56.5 & 59.0 & 15.1 & \underline{10.6} & 63.2 & \underline{18.5} & 26.8 \\
    Moon & \underline{57.5} & 54.1 & 74.5 & 56.0 & 60.5 & 15.5 & 14.5 & 61.9 & 20.1 & 28.0 \\
    FedMLB & 54.0 & 50.0 & 72.9 & \underline{57.6} & 58.6 & 16.3 & 12.2 & 72.1 & 23.0 & 30.9 \\ \hline
    FedDA-cyclic & 56.3 & \textbf{59.1} & \textbf{83.9} & \textbf{61.1} & \textbf{65.1} & 14.8 & 14.5 & \textbf{43.8} & 22.8 & \textbf{24.0}\\
    FedDA-joint & \textbf{60.9} & \underline{54.2} & \underline{80.8} & 53.8 & \underline{62.4} & \textbf{11.9} & 12.3 & 53.3 & 22.8 & \underline{25.1} \\ \hline
        \end{tabular}}
    \caption{The performance comparison of FedDA and other state-of-the-art methods on the two-client setting for the two medical image segmentation tasks. The bold value and underline value denote the best and the second-best performance in each metric, respectively.}
    \label{2client_c}
\end{table}
\section{Experiments And Results}
\subsection{Datasets and Evaluation Metrics}
To validate the proposed FedDA framework, we conducted experiments on two medical image segmentation tasks: cardiac substructure segmentation and abdominal multi-organ segmentation. For cardiac substructure segmentation, we used the MM-WHS Challenge 2017 dataset~\citep{zhuang2016multi} targeting five cardiac substructures (myocardium (MYO), right atrium (RA), right ventricle (RV), left atrium (LA), and left ventricle (LV)). For abdominal multi-organ segmentation, we used CT images from the MICCAI 2015 Challenge and MRI images from the ISBI 2019 CHAOS Challenge~\citep{kavur2021chaos}, targeting four organs (liver (LR), right kidney (RK), left kidney (LK), and spleen (SP)). 
We first separately divided the training and testing sets for the CT and MRI modalities at a ratio of 4:1. The test sets from the different modalities are then combined to form a unified \textbf{global test set}, which is used to comprehensively evaluate the generalization capability of the model across all modalities. 
\begin{table}[]
    \centering
    \fontsize{9pt}{10pt}\selectfont
    \setlength{\tabcolsep}{0.5mm} 
    \scalebox{0.752}{
        \begin{tabular}{c|cccccc|cccccc}
        \hline
    \multirow{3}{*}{Methods} & \multicolumn{12}{c}{Cardiac Substructure Segmentation (MMWHS)} \\ \cline{2-13}    
    & \multicolumn{6}{c}{Dice (\%)} & \multicolumn{6}{c}{HD95} \\ \cline{2-13} 
    & MYO & RA & RV & LA & LV & Mean & MYO & RA & RV & LA & LV & Mean\\ \hline  
    FedAvg & 60.6 & 82.8 & 85.8 & 80.7 & 85.7 & 79.1 & 28.5 & 30.6 & 25.8 & 28.9 & 23.7 & 27.5 \\
    Krum & 52.9 & 72.8 & 67.9 & 70.4 & 73.9 & 67.6 & 57.3 & 46.6 & 47.1 & 37.8 & 37.5 & 45.3 \\
    Fedprox & 58.6 & 83.1 & 85.1 & 84.0 & 86.5 & 79.5 & 27.1 & 29.4 & 25.8 & 22.7 & 21.0 & 25.2 \\
    FedBN & 62.0 & 83.5 & 85.6 & 83.9 & 86.3 & 80.3 & 28.2 & 31.2 & \textbf{23.1} & 22.6 & 22.4 & 25.5 \\
    Moon & 62.6 & \underline{84.4} & \underline{86.9} & \underline{85.5} & 87.3 & 81.3 & 26.6 & \underline{27.6} & \underline{23.7} & 20.1 & 19.8 & 23.5 \\
    FedMLB & 58.9 & 83.1 & 85.7 & 85.2 & 84.8 & 79.5 & 29.7 & 31.5 & 24.7 & \underline{19.9} & 25.2 & 26.2 \\ \hline
    FedDA-cyclic & \underline{63.6} & \textbf{85.1} & 86.9 & 84.8 & \underline{88.2} & \underline{81.7} & \textbf{25.9} & \textbf{25.6} & 24.2 & 21.2 & \textbf{18.1} & \textbf{23.0} \\
    FedDA-joint & \textbf{65.1} & 83.9 & \textbf{88.2} & \textbf{86.5} & \textbf{88.7} & \textbf{82.5} & \underline{26.3} & 28.62 & 24.5 & \textbf{17.3} & \underline{18.8} & \underline{23.1} \\ \hline
        \end{tabular}}
    \caption{The objective evaluation of FedDA and other state-of-the-art methods on the three-client setting for Cardiac Substructure Segmentation. The bold value and underline value denote the best and the second-best performance in each metric, respectively.}
    \label{3client}
\end{table}
\par For the tasks of cardiac substructure segmentation and abdominal multi-organ segmentation, we initially conducted experiments with two clients. While maintaining the composition of the unified global test set, we distributed the training sets from different modalities to separate clients to simulate the federated learning environment. Subsequently, we extended the experiments on cardiac substructure segmentation to a three-client setting. Building upon the original two-client configuration, we partitioned the MRI training set into two sub-training sets in a 1:1 ratio, assigning each subset to two clients as their local data while the client holding the CT data remained unchanged. Last, we conducted a four-client simulation experiment on the abdominal multi-organ segmentation task. We partitioned the CT and MRI training sets into four sub-training sets each in a 1:1 ratio and assigned these subsets as the local datasets for the four clients to simulate the federated learning environment. All images are resized to 512$\times$512 pixels for the experiment. \textbf{All datasets are split based on patient volumes to avoid patient overlapping between training and testing sets as well as between different clients.} For evaluation, we adopt two commonly-used metrics of Dice coefficient (Dice) and Hausdorff distance (HD95), to quantitatively evaluate the segmentation results on the whole object region and the surface shape respectively.
\subsection{Implementation Details}
We implement all methods using three frameworks: PyTorch~\citep{paszke2019pytorch}, MMsegmentation~\citep{mmseg2020}, and Flower~\citep{beutel2020flower}. We select SegFormer~\citep{xie2021segformer} as the model for experiments. All experiments were conducted on a NVIDIA RTX 4090 GPU. We utilize Adam optimizer with learning rate of 1e-3 for backbone and learning rate of 1e-6 for discriminator and a weight decay of 1e-5. For the federated learning simulation, we conduct experiments over a total of 100 communication rounds. Each client
performs local training for 1 epoch in each round. The weight decay for both the local models and the discriminator is set to 0.01. The segmentation loss is computed using the cross-entropy loss while
the discriminator loss is calculated using the binary cross entropy loss.
\subsection{Comparison Methods}
In our experiments, we employed six notable strategies: (1) FedAvg (AISTATS17), (2) Krum (NeurIPS17), (3) FedProx (MLSys20), (4) FedBN (ICML2021), (5) Moon (CVPR2021), (6) FedMLB (ICML2022). We conducted comprehensive evaluations by comparing the results against our own baseline methods.
\subsection{Experiments Results}
We conduct experiments in multiple federated learning scenarios on cardiac substructure segmentation and abdominal multi-organ segmentation tasks and compare the objective segmentation performance of our method and SOTA approaches in Tab.~\ref{2client_m}, Tab.~\ref{2client_c}, Tab.~\ref{3client} and Tab.~\ref{4client}. 
Tab.~\ref{2client_m} and Tab.~\ref{2client_c} present the quantitative results for the two-client experiments on cardiac substructure segmentation and abdominal multi-organ segmentation tasks. Tab.~\ref{3client} and Tab.~\ref{4client} present the quantitative results of the cardiac substructure segmentation task in the 3-client scenario and the abdominal multi-organ segmentation task in the 4-client scenario. Compared with these methods, both FedDA-cyclic and FedDA-joint achieve higher overall performance and demonstrate good performance in most structural segmentation tasks. This is primarily due to the adversarial training mechanism, which can effectively reduce the differences between different models. After incorporating feature-level adversarial training, the model not only achieves a substantial enhancement in performance but also reduces its bias towards specific modalities.
\begin{table}[]
    \centering
    \fontsize{9pt}{10pt}\selectfont
    \setlength{\tabcolsep}{0.8mm} 
    \scalebox{0.825}{
        \begin{tabular}{c|ccccc|ccccc}
        \hline
    \multirow{3}{*}{Methods} & \multicolumn{10}{c}{Abdominal Multi-Organ Segmentation (CHAOS)} \\ \cline{2-11}    
    & \multicolumn{5}{c}{Dice (\%)} & \multicolumn{5}{c}{HD95} \\ \cline{2-11} 
    & LK & SP & LR & RK & Mean & LK & SP & LR & RK & Mean \\ \hline    
    FedAvg & \textbf{55.5} & 32.1 & 67.7 & 53.3 & 52.1 & 15.5 & 14.6 & 95.2 & 30.9 & 39.1 \\
    Krum  & 49.6 & 20.5 & 75.0 & 51.3 & 49.1 & 15.3 & 12.6 & 62.2 & 19.9 & 27.5 \\
    Fedprox & 49.7 & 46.9 & 70.2 & 53.2 & 55.0 & 15.7 & 12.2 & 61.2 & 20.9 & 27.5 \\
    FedBN & \underline{54.4} & 46.4 & 72.0 & 53.3 & 56.5 & 16.4 & 11.1 & 57.2 & \textbf{18.8} & 25.9 \\
    Moon & 49.6 & 41.1 & 74.4 & 55.4 & 55.1 & 20.0 & \underline{10.7} & 64.6 & 25.3 & 30.2 \\
    FedMLB & 52.3 & 32.5 & 75.1 & 53.8 & 53.4 & \textbf{12.8} & \textbf{8.9} & 72.5 & \underline{18.9} & 28.3 \\ \hline
    FedDA-cyclic & 53.3 & \underline{54.3} & \textbf{76.7} & \underline{55.7} & \underline{60.1} & \underline{14.9} & 12.2 & \textbf{52.4} & 23.6 & \textbf{25.8} \\
    FedDA-joint & 53.9 & \textbf{57.8} & \underline{76.7} & \textbf{57.4} & \textbf{61.5} & \underline{14.9} & 12.1 & \underline{53.5} & 23.1 & \underline{25.9} \\ \hline
        \end{tabular}}
    \caption{The objective evaluation of FedDA and other state-of-the-art methods on the four-client setting for Abdominal Multi-Organ Segmentation. The bold value and underline value denote the best and the second-best performance in each metric, respectively.}
    \label{4client}
\end{table}
\begin{figure*}
    \centering
    \includegraphics[width=\linewidth]{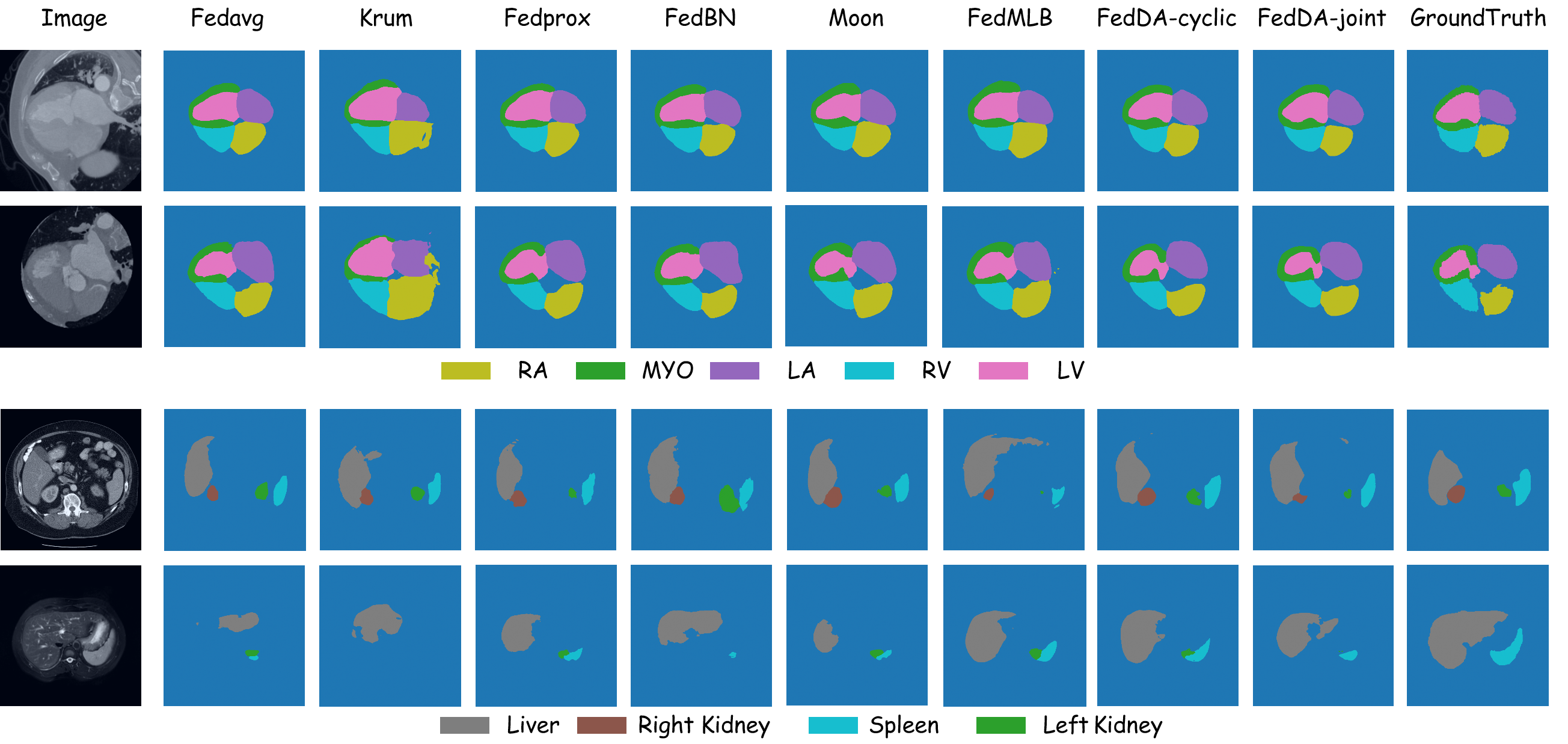}
    \caption{Qualitative comparison on the segmentation results of different methods on cardiac substructure segmentation and abdominal multi-organ segmentation tasks in the two-client scenario.}
    \label{vis}
\end{figure*}
In the cardiac substructure segmentation task, our methods FedDA-cyclic and FedDA-joint achieved certain performance enhancements compared to previous federated algorithms, with improvements of 2.63\% in Dice score and 4.3 in HD in the two-client scenario (Tab.~\ref{2client_m}), and 1.1\% in Dice score and 0.49 in HD in the three-client scenario (Tab.~\ref{3client}). Moreover, the performance of the two FedDA variants we proposed consistently ranked in the top two positions in the majority of substructures segmentation. The FedDA-cyclic and FedDA-joint are meticulously designed to leverage the principles of domain adaptation within the framework of federated learning. FedDA-cyclic operates by cyclically adapting models across different clients and domains, enabling a continuous exchange of knowledge that gradually aligns the local models with the global optimal solution. On the other hand, FedDA-joint takes a more holistic approach, jointly optimizing the model training and domain adaptation processes, which allows for a more coordinated and efficient utilization of data from diverse sources. For the abdominal multi-organ segmentation tasks, our proposed FedDA-cyclic and FedDA-joint also achieve significant performance improvements for dice from 66.75\% to 68.23\% and HD from 26.28 to 24.75 in two-client scenario (Tab.~\ref{2client_c}) and for dice from 64.25\% to 67.06\% and HD from 27.56 to 25.85 in four-client scenario (Tab.~\ref{4client}). The consistent performance improvements across both cardiac and abdominal segmentation tasks validate the generalizability and efficiency of the FedDA framework in addressing the challenges of medical image segmentation in federated learning environments.
\begin{table}[]    
\centering    
\fontsize{9pt}{10pt}\selectfont    
\setlength{\tabcolsep}{0.5mm}     
\scalebox{0.785}{        
\begin{tabular}{c|cccccc|cccccc}        
\hline    
\multirow{3}{*}{Methods} & \multicolumn{12}{c}{Cardiac Substructure Segmentation (MMWHS)} \\ \cline{2-13}    
& \multicolumn{6}{c}{Dice (\%) $\uparrow$} & \multicolumn{6}{c}{HD95$\downarrow$} \\ \cline{2-13}    
& MYO & LA & LV & RA & RV  & Mean & MYO & LA & LV & RA & RV & Mean\\ \hline    
FedAvg & 61.2 & 83.8 & 87.2 & 82.7 & 86.3 & 80.2 & 26.8 & 28.7 & 23.3 & 24.0 & 20.8 & 24.7 \\    
FedAvg-c & 67.6 & 88.1 & 88.8 & 88.3 & 88.9 & 84.3 & 21.7 & 25.0 & 20.4 & 16.5 & 17.7 & 20.3 \\    
FedAvg-j & \textbf{68.2} & \textbf{88.3} & \textbf{88.9} & \textbf{88.8} & \textbf{89.7} & \textbf{84.8} & \textbf{21.3} & \textbf{22.4} & \textbf{20.4} & \textbf{15.3} & \textbf{16.9} & \textbf{19.3} \\ \hline    
Krum & 47.9 & \textbf{74.8} & 79.2 & 66.8 & \textbf{85.0} & 70.8 & 41.5 & \textbf{38.3} & 41.4 & 60.9 & \textbf{22.3} & 40.9\\    
Krum-c & 48.1 & 73.2 & 78.1 & \textbf{75.4} & 77.1 & 70.4 & \textbf{37.2} & 39.6 & 34.9 & \textbf{27.9} & 29.2 & \textbf{33.8}\\    
Krum-j & \textbf{54.6} & 73.7 & \textbf{79.6} & 73.1 & 79.5 & \textbf{72.1} & 38.6 & 42.3 & \textbf{33.1} & 31.7 & 30.2 & 35.2\\    
\hline        
\end{tabular}}    
\caption{The performance comparison of incorporating feature-level mechanisms in a two-client setting for two medical image segmentation tasks. Krum-c refers to the integration of FedDA-cyclic with the Krum aggregation algorithm, while FedAvg-j signifies the combination of FedDA-joint with the FedAvg aggregation algorithm.}    
\label{exten_m}
\end{table}
\begin{figure*}    
\centering    
\includegraphics[width=\linewidth]{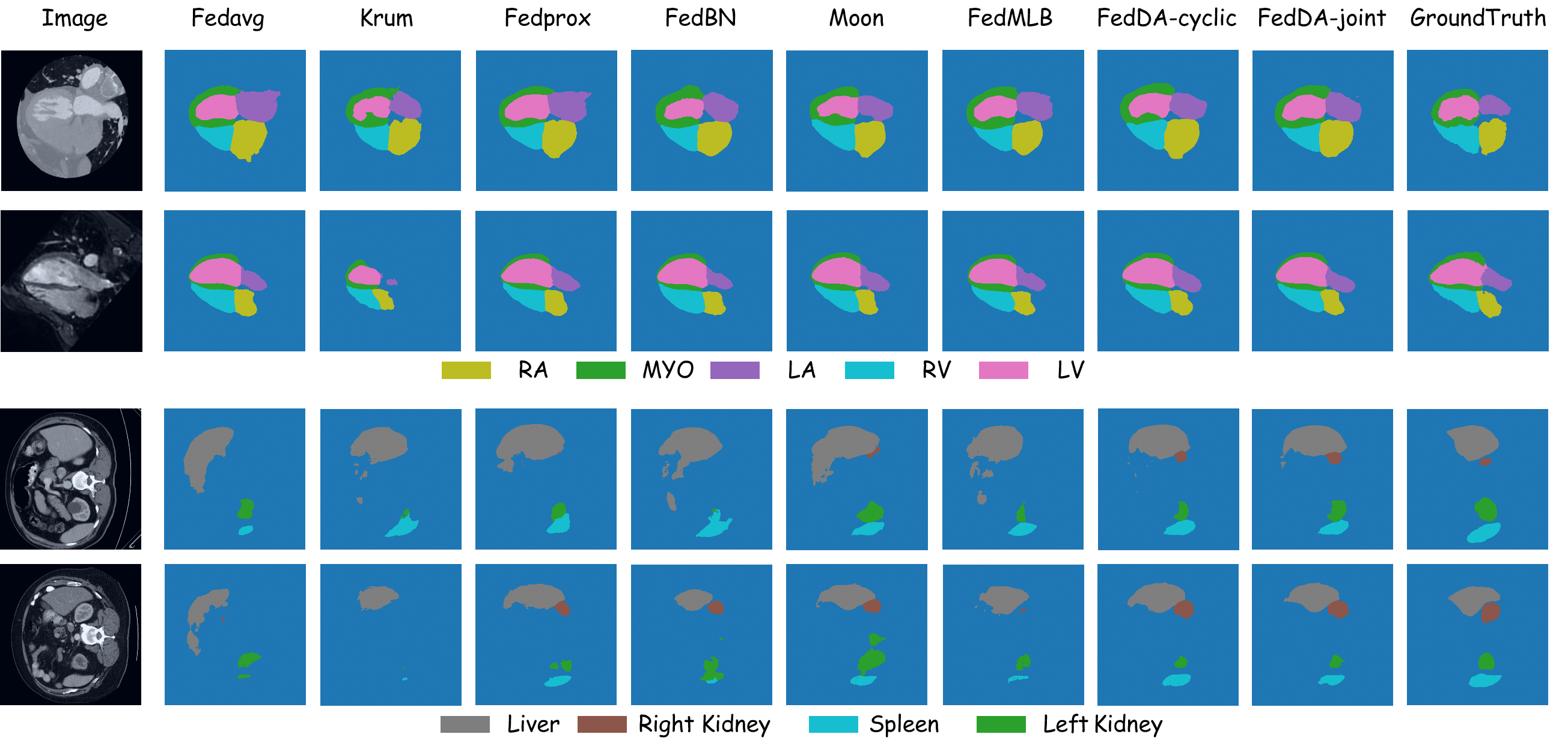}    
\caption{Qualitative comparison on the segmentation results of different methods on two medical image segmentation task. The first two rows show the result of Cardiac Substructure Segmentation in three-client scenario while the last two rows show the result of Abdominal Multi-Organ Segmentation in four-client scenario.}    
\label{appvis}
\end{figure*}
\subsection{Visualization}
\begin{table}[]    
\centering    
\fontsize{9pt}{10pt}\selectfont    
\setlength{\tabcolsep}{0.8mm}     
\scalebox{0.88}{        
\begin{tabular}{c|ccccc|ccccc}        
\hline    
\multirow{3}{*}{Methods} & \multicolumn{10}{c}{Abdominal Multi-Organ Segmentation (CHAOS)}\\ \cline{2-11}    
& \multicolumn{5}{c}{Dice (\%) $\uparrow$} & \multicolumn{5}{c}{HD95$\downarrow$} \\ \cline{2-11}    
& LK & SP & LR & RK & Mean & LK & SP & LR & RK & Mean\\ \hline    
FedAvg & 54.6 & 44.1 & 72.6 & 57.1 & 57.1 & 14.9 & 9.9 & 69.7 & 22.5 & 29.2 \\    
FedAvg-c & 56.3 & \textbf{59.1} & \textbf{83.9} & \textbf{61.1} & \textbf{65.1} & 14.8 & 14.5 & \textbf{43.8} & \textbf{22.8} & \textbf{24.0}\\    
FedAvg-j & \textbf{60.9} & 54.2 & 80.8 & 53.8 & 62.4 & \textbf{11.9} & \textbf{12.3} & 53.3 & 22.8 & 25.1 \\ \hline    
Krum & \textbf{53.1} & 50.7 & 71.2 & \textbf{49.7} & 56.2 & 13.0 & 10.9 & 78.7 & 15.7 & 29.6\\    
Krum-c & 57.3 & 52.1 & 80.2 & \textbf{63.2} & \textbf{63.2} & 15.8 & 9.19 & \textbf{54.6} & 22.7 & \textbf{25.6}\\    
Krum-j & 54.1 & \textbf{47.1} & \textbf{70.4} & 54.8 & 56.6 & \textbf{10.6}& \textbf{10.6} & 75.3 & \textbf{15.4} & 27.9 \\    
\hline        
\end{tabular}}    
\caption{The performance comparison of incorporating feature-level mechanisms in a two-client setting for two medical image segmentation tasks. Krum-c refers to the integration of FedDA-cyclic with the Krum aggregation algorithm, while FedAvg-j signifies the combination of FedDA-joint with the FedAvg aggregation algorithm.}    
\label{exten_c}
\end{table}
As shown in Fig.~\ref{vis}, we conducted visualization experiments in a two-client federated scenario for the two medical segmentation tasks. It is evident that our proposed approach effectively augments the generalization capacity of the model across diverse clients and modalities. Additionally, we conducted comprehensive visualization analysis for scenarios involving three and four clients, respectively. The results of these analyses are depicted in Fig.~\ref{appvis}. In Fig.~\ref{appvis}, the first two rows illustrate the segmentation results for the Cardiac Substructure Segmentation task in a three-client scenario, while the last two rows show the results of the Abdominal Multi-Organ Segmentation task in a four-client scenario. The consistent improvements observed across various metrics and scenarios highlight the reliability of our methods in enhancing segmentation accuracy and model generalization. Specifically, the visualizations reveal that our approach effectively reduces segmentation errors and improves boundary delineation, even in the presence of complex sub-structures and varying data distributions. The consistent improvements observed across various metrics and scenarios, from the two-client to the four-client cases, strongly highlight the reliability of our methods in enhancing segmentation accuracy and model generalization. Specifically, upon a closer examination of the visualizations, it becomes evident that our approach effectively reduces segmentation errors, even in areas with intricate sub-structures. Overall, the consistent performance gains across all tasks in different client scenarios, validate the robustness and adaptability of our framework.
\begin{table}[]    
\centering    
\fontsize{9pt}{10pt}\selectfont    
\setlength{\tabcolsep}{0.52mm}         
\scalebox{0.752}{            
    \begin{tabular}{c|cccccc|cccccc}            
    \hline        
    \multirow{3}{*}{Methods} & \multicolumn{12}{c}{Cardiac Substructure Segmentation (MMWHS)} \\ \cline{2-13}        
    & \multicolumn{6}{c}{Dice (\%) $\uparrow$} & \multicolumn{6}{c}{HD95$\downarrow$}  \\ \cline{2-13}        
    & MYO & LA & LV & RA & RV  & Mean & MYO & LA & LV & RA & RV & Mean\\ \hline        
    FedAvg & 61.2 & 83.8 & 87.2 & 82.7 & 86.3 & 80.2 & 26.8 & 28.7 & 23.3 & 24.0 & 20.8 & 24.7 \\        
    FedDA-cyc1 & 66.5 & 84.2 & 88.1 & 84.3 & 87.3 & 82.1 & 29.3 & 30.1 & 23.1 & 23.7 & 21.2 & 25.5\\        
    FedDA-cyc2 & 63.6 & 86.2 & 87.8 & 85.1 & 87.4 & 82.0 & 27.4 & 27.3 & 22.1 & 21.9 & 18.8 & 23.5\\        
    FedDA-joint1 & 66.1 & 84.3 & 87.1 & 85.6 & 87.5 & 82.1 & 24.2 & 26.7 & 25.5 & 19.7 & 19.1 & 23.1\\    
    FedDA-joint2 & 63.6 & 83.9 & 87.7 & 85.9 & 87.9 & 81.8 & 24.7 & 28.1 & 23.6 & 21.1 & 18.1 & 23.1\\ \hline 
    FedDA-cyclic & \underline{67.6} & \underline{88.1} & 88.8 & \underline{88.3} & 88.9 & \underline{84.3} & \underline{21.7} & \underline{25.0} & \textbf{20.4} & \underline{16.5} & \underline{17.7} & \underline{20.3}\\  
    FedDA-joint & \textbf{68.2} & \textbf{88.3} & \textbf{88.9} & \textbf{88.8} & \textbf{89.7} & \textbf{84.8} & \textbf{21.3} & \textbf{22.4} & \textbf{20.4} & \textbf{15.3} & \textbf{16.9} & \textbf{19.3}\\ \hline            
    \end{tabular}
}    
\caption{Ablation results to analyze the effect of the feature-level adversarial training in our method. FedDA-cyc1 means adversarial training conducted only by Client1 on FedDA-cyclic and FedDA-joint2 means adversarial training conducted only by Client2 on FedDA-joint. The bold value denote the best performance in each metric. }    
\label{ablation_mmwhs}
\end{table}
\subsection{Extensibility With Other Methods}
FedDA is a federated learning strategy that relies on local training and employs feature-level adversarial mechanisms to align model representations across different clients. Moreover, at the level of model parameter aggregation, FedDA exhibits seamless compatibility with a broad spectrum of existing algorithms, including Krum~\citep{blanchard2017machine} and FedAvg~\citep{mcmahan2017communication}, which are predicated on model parameter aggregation. Thus, we compare the performance on two federated learning strategies as well as their performance after incorporating feature-level adversarial mechanism. As shown in Tab.~\ref{exten_m} and Tab.~\ref{exten_c}, the incorporation of feature-level adversarial mechanism leads to significant performance improvements for two segmentation tasks in two-client scenario. Basically, for each federated strategy and each segmentation category, the performance is improved by incorporating adversarial learning. As shown in Tab.~\ref{exten_m} and Tab.~\ref{exten_c}, it is obvious that adversarial training also leads to improvements for abdominal segmentation in various metrics. 
\begin{table}[]
    \centering
    \fontsize{9pt}{10pt}\selectfont
    \setlength{\tabcolsep}{0.5mm} 
    \scalebox{0.89}{
        \begin{tabular}{c|ccccc|ccccc}
        \hline
    \multirow{3}{*}{Methods} & \multicolumn{10}{c}{Abdominal Multi-Organ Segmentation (CHAOS)} \\ \cline{2-11}    
    & \multicolumn{5}{c}{Dice (\%) $\uparrow$} & \multicolumn{5}{c}{HD95 $\downarrow$} \\ \cline{2-11} 
    & LK & SP & LR & RK & Mean & LK & SP & LR & RK & Mean \\ \hline    
    FedAvg & 54.6 & 44.1 & 72.6 & 57.1 & 57.1 & 14.9 & \textbf{9.9} & 69.7 & 22.5 & 29.2 \\
    FedDA-cyc1  & 56.2 & 55.7 & 74.8 & 56.1 & 60.7 & 14.6 & 14.0 & 56.3 & \textbf{15.7} & 25.4 \\
    FedDA-cyc2 & 53.9 & 51.2 & 81.0 & 57.5 & 60.9 & 17.9 & 14.1 & 49.7 & 23.6 & 26.3 \\ 
    FedDA-joint1  & 57.6 & 51.0 & 79.7 & 50.8 & 59.8 & 14.5 & 12.2 & 55.3 & 22.6 & 26.1 \\
    FedDA-joint2 & 56.2 & 52.6 & 75.5 & 55.2 & 59.9 & 16.2 & 13.0 & 56.6 & 19.0 & 26.2 \\ \hline 
    FedDA-cyclic & 56.3 & \textbf{59.1} & \textbf{83.9} & \textbf{61.1} & \textbf{65.1} & 14.8 & 14.5 & \textbf{43.8} & 22.8 & \textbf{24.0} \\
    FedDA-joint & \textbf{60.8} & 54.2 & 80.8 & 53.8 & 62.4 & \textbf{11.9} & 12.3 & 53.3 & 22.8 & 25.1 \\ \hline
        \end{tabular}}
    \caption{Ablation results to analyze the effect of the feature-level adversarial training in our method. FedDA-cyc1 means adversarial training conducted only by Client1 on FedDA-cyclic and FedDA-joint2 means adversarial training conducted only by Client2 on FedDA-joint. The bold value denote the best performance in each metric.}
    \label{ablation}
\end{table}
\par After incorporating adversarial training, the model not only achieves a substantial enhancement in performance also reduces its bias towards modalities. It is evident that the introduction of adversarial learning significantly improves the model's generalization ability. In terms of quantitative metrics, the mean Dice score (mDice) has increased significantly by approximately 2.3\% and the Hausdorff Distance (HD) has decreased by 4.6 on average after the implementation of adversarial training, indicating improved segmentation accuracy and enhanced ability to delineate complex structures. It is evident that the introduction of adversarial learning significantly improves the model’s generalization ability. Without adversarial training, the model might learn to rely on modality-specific cues, leading to poor performance when encountering images from an unseen modality. However, with the addition of adversarial training, the model is compelled to extract features that are invariant to modality differences, enabling it to perform consistently well across different imaging types.
\begin{figure}
    \centering
    \includegraphics[width=\linewidth]{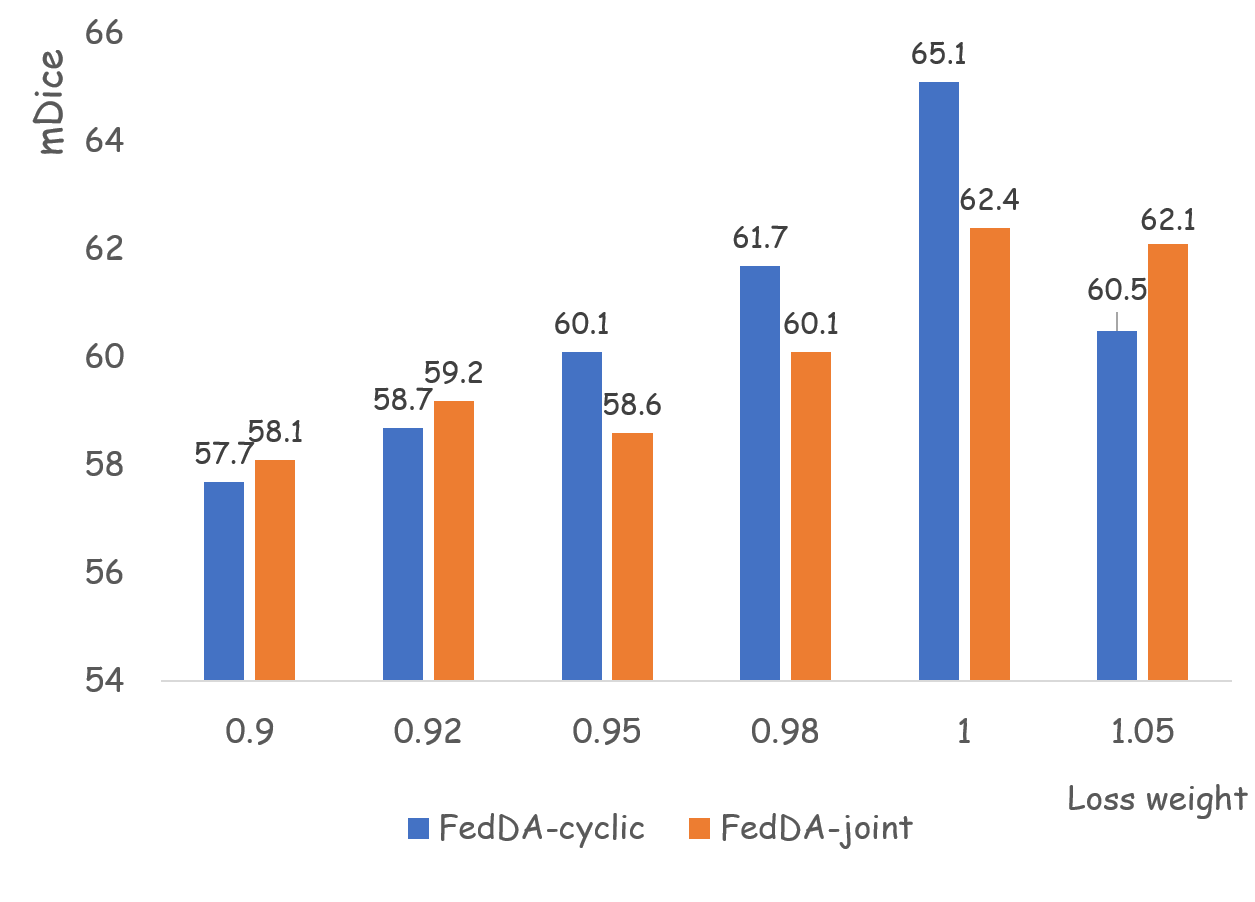}
    \caption{Comparison of mDice scores for segmentation results with different discriminator loss weights in Abdominal Multi-Organ Segmentation in two-client scenario.}
    \label{hyper_dice}
\end{figure}
\subsection{Hyperparameter Experimentation in FedDA}
For Abdominal Multi-Organ Segmentation task, we tune the loss weight for the discriminator on FedDA-cyclic and FedDA-joint for two-client experiments on this task. In each comparison, we fix the total number of global communication rounds, the proportion of clients uploaded in each round, and the number of local training epochs. The experimental results are visually presented below, from which we can distinctly observe that the segmentation performance varies significantly with different discriminator loss weights. 
As shown in Fig.~\ref{hyper_dice}, the segmentation performance metric mean Dice (mDice), initially increases and then decreases with the rise in the discriminator loss weight. Fig.~\ref{hyper_dice} illustrates the variation of the Hausdorff Distance (HD) for model segmentation with the increase in loss weight. It is evident from the figure that the HD initially decreases and then increases as the loss weight rises. This suggests that the segmentation accuracy of the model improves with a moderate increase in loss weight, but deteriorates when the loss weight exceeds a certain threshold. Specifically, an optimal loss weight can enhance the model’s ability to distinguish between different organs, leading to higher segmentation accuracy. When the loss weight becomes excessively large, it may cause overfitting to the training data, resulting in a decline in generalization performance.
\begin{figure}
    \centering
    \includegraphics[width=\linewidth]{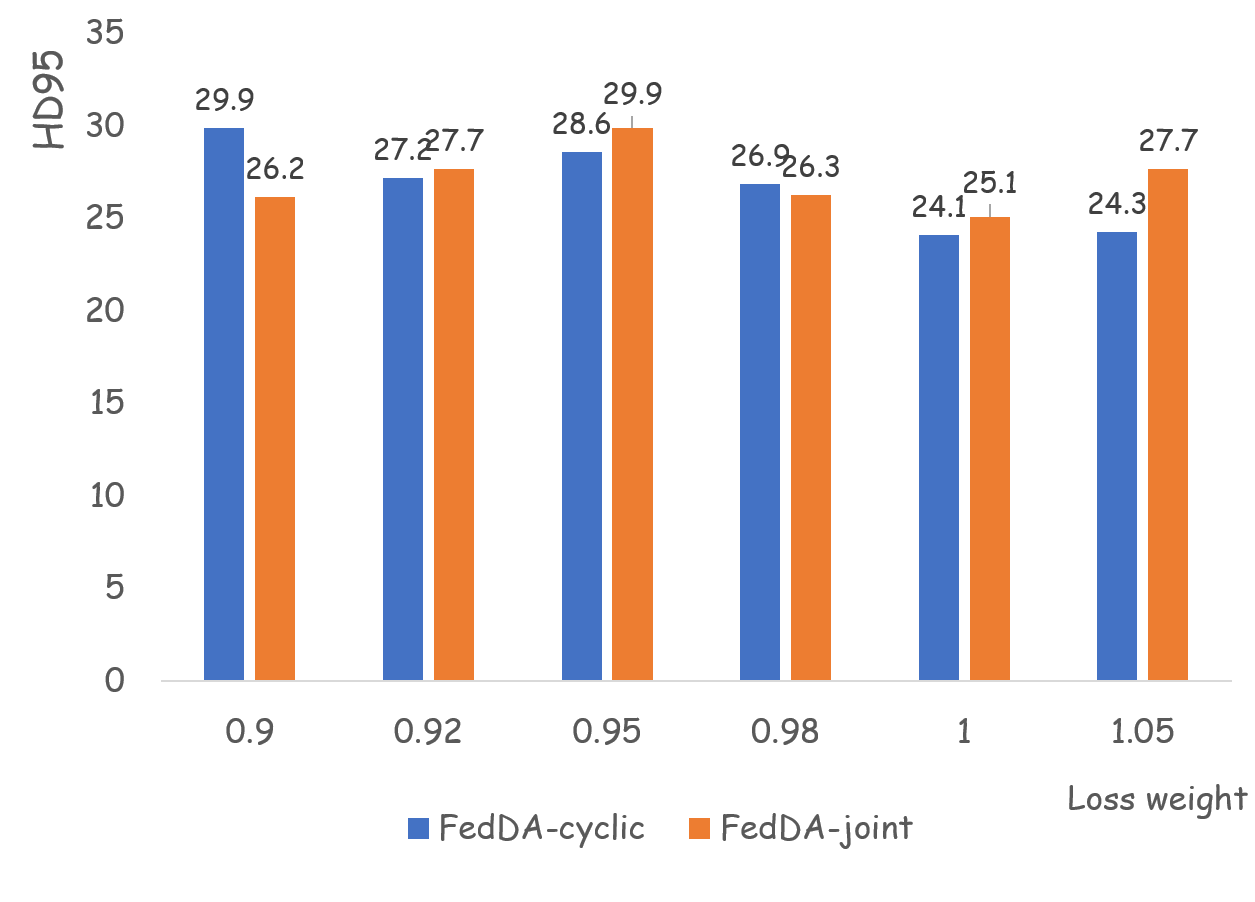}
    \caption{Comparison of HD95 scores for segmentation results with different discriminator loss weights in Abdominal Multi-Organ Segmentation in two-client scenario.}
    \label{hyper_dice}
\end{figure}
\subsection{Ablation Study}
The proposed FedDA consists of FedDA-cyclic and FedDA-joint, depending on how to get the target feature maps from the perspective of each client. Both of them achieve an obvious improvement over the FedAvg baseline by incorporating feature-level adversarial training. Therefore, we conduct seven sets of experiments to ensure the effectiveness of each component in two-client scenario based on abdominal multi-organ segmentation tasks: (1) the baseline FedAvg, (2) feature-level adversarial training conducted only by Client1 or (3) Client2 on FedDA-cyclic, (4) feature-level adversarial training conducted only by Client1 or (5) Client2 on FedDA-joint, (6) FedDA-cyclic and (7) FedDA-joint. As shown in Tab.~\ref{ablation}, we find that FedAvg can derive benefits from adversarial settings. This phenomenon demonstrates the effectiveness of our method for multi-modality federated segmentation. A single client conducting feature-level adversarial learning can still achieve performance improvements compared to the federated learning baseline FedAvg. This indicates that the feature-level adversarial learning strategy we introduced successfully mitigates the differences between models, thereby enhancing overall consistency and performance. When adversarial training are conducted between the two clients, optimal performance is typically achieved. This suggests that the interactive adversarial mechanism fosters a more robust feature alignment process, where models from different clients can mutually constrain and optimize each other's representations. Such a collaborative framework not only narrows the gap between local models but also promotes the learning of more generalized and discriminative features for multi-modality segmentation tasks.
\section{Discussion}
\subsection{Representation Privacy Analysis}
FedDA is capable of extracting and disseminating representation information across clients, which consequently enables it to effectively alleviate the impact of the non-IID data distribution phenomenon. Nevertheless, the transmission of feature representations may raise concerns regarding potential privacy leakage. In practice, the process of encoding a grayscale medical image \( I \in \mathbb{R}^{H \times W} \) into a feature tensor \( F \in \mathbb{R}^{C' \times H' \times W'} \) via the backbone is a high-dimensional and complex transformation. 
\par Firstly, high-dimensional feature maps are complex tensors that lack intuitive interpret ability, and their inherent content does not directly reveal patient privacy. Feature maps do not explicitly reveal pathological lesion locations, shapes, sizes, or personal identifiers such as names, which could be directly discerned from raw local data images.
\par Recent studies~\citep{liang2021swinir,zhou1988image,zamir2021multi} often employ transposed convolution and up sampling techniques to reconstruct original data from feature maps. However, these approaches typically require a substantial amount of raw data as labels for each training session. Nonetheless, within the scenario of federated learning, the non-sharing of raw data across different clients impedes the efficacy of these conventional methods. Even though there are now invertible neural networks~\citep{ardizzone2018analyzing,mao2024denoising} used for data recovery or data repair, its core mathematical formulations (Equ.~\ref{inn}) of these networks limit their ability to recover the original data from feature maps.
\begin{equation}
[\mathbf{y}, \mathbf{z}] = f(\mathbf{x}; \theta) = [f_{\mathbf{y}}(\mathbf{x}; \theta), f_{\mathbf{z}}(\mathbf{x}; \theta)] = g^{-1}(\mathbf{x}; \theta) \quad\\
\label{inn}
\end{equation}$[\mathbf{y}, \mathbf{z}]$: Represents the output obtained after the input data $\mathbf{x}$ is processed by the forward function $f$ of the network, where $\mathbf{y}$ and $\mathbf{z}$ are two parts of the output. $g^{-1}(\mathbf{x}; \theta)$: represents the inverse function. This shows that an invertible neural network must maintain consistency between the dimensions of the input and output. Nevertheless, the spatial dimensions of the feature maps typically deviate from those of the original input data. Then, the training process necessitates the use of original images as inputs to facilitate the recovery of the initial data. Similarly, in our proposed approach, the original raw images remains stored locally on each data client, and it does not lead to the leakage of local raw images.
\begin{equation}
\begin{aligned}
& q(\mathbf{x} = g(\mathbf{y}, \mathbf{z}; \theta) \mid \mathbf{y}) = p(\mathbf{z}) \left| J_{\mathbf{x}} \right|^{-1} & \\
& J_{\mathbf{x}} = \det \left( \frac{\partial g(\mathbf{y}, \mathbf{z}; \theta)}{\partial [\mathbf{y}, \mathbf{z}]} \bigg|_{\mathbf{y}, f_{\mathbf{z}}(\mathbf{x})} \right) &
\end{aligned}
\label{inncore}
\end{equation}\( |J_{\mathbf{x}}| \) is the determinant of the Jacobian matrix of the transformation \( g \) at \( \mathbf{x} \). Actually, the transformation of an image into a feature map represents a process where two-dimensional data is converted into a three-dimensional representation. However, the reconstruction of the original image from the feature map involves mapping from a higher-dimensional space to a lower-dimensional space, which is not unique. \par Indeed, there are precedents for exchanging information beyond model parameters among various clients. For instance, the FedST~\citep{ma2024fedst} alleviates the impact of non-IID data by transferring style information from original images. Therefore, transferring feature maps among clients has little impact on the privacy of clients which is not considered a violation of the fundamental principles of federated learning.
\subsection{Communication Cost Analysis}
Communication cost is a crucial factor in federated learning, especially when dealing with large-scale distributed systems where bandwidth and latency can significantly impact the efficiency and scalability of the training process. Owing to the distinct mechanisms of feature map transmission, the communication burdens associated with FedDA-cyclic and FedDA-joint are different. Compared with other federated algorithms such as FedAvg~\citep{mcmahan2017communication} or Krum~\citep{blanchard2017machine}, FedDA-joint do not incur additional communication burden, as the parameters of the local discriminators are not involved in the aggregation process. Due to the uploading of local feature maps to the server, the communication burden of FedDA-cyclic is slightly higher. The transmission of feature tensors \( F \in {R}^{C' \times H' \times W'} \) typically incurs communication costs on the order of tens of thousands. However, most contemporary segmentation models possess millions of parameters. In comparison, the communication burden associated with the transmission of feature tensors could be negligible. Therefore, while the transmission of feature maps in FedDA-cyclic does introduce additional communication overhead, it remains within an acceptable range.
\section{Conclusion}
We present a practical solution to improve the performance under a non-typical non-IID federated scenario, where some clients possess medical images of one modality, while other clients hold images of a different modality. Specifically, our solution improves the model's generalization performance on each client and modality by conducting adversarial training between target and source feature maps during local training iterations. Depending on the source of the target feature map, two different modes are derived: FedDA-cyclic and FedDA-joint. Subsequently, we conduct rigorous experiments on three medical datasets in different experiment settings. The experimental results validate the effectiveness of FedDA which demonstrates a significant improvement on medical segmentation in objective and subjective assessment. Moreover, our experiments also demonstrate that the proposed method can be effectively integrated into some aggregation-based federated learning algorithms to jointly enhance experimental performance. The proposed learning framework for encouraging feature-level adversarial training is also generally extendable to other collaborative problems.

\bibliographystyle{model2-names} 
\bibliography{aaai2026}
\end{document}